\documentclass[11pt,a4paper]{article}
\usepackage[hyperref]{acl2020}
\usepackage{times}
\usepackage{latexsym}
\usepackage{amsmath}
\usepackage{amsfonts}
\usepackage{graphicx}
\usepackage{booktabs}
\usepackage{microtype}
\usepackage{textcomp}
\usepackage{ulem}

\aclfinalcopy 

\newcommand{\bmat}[1]{\text{\textbf{#1}}}
\newcommand{\bvec}[1]{\boldsymbol{#1}}
\newcommand{\floor}[1]{\lfloor #1 \rfloor}
\newcommand\base{\textsc{base}}
\newcommand\hcsa{\textsc{hc-sa}}
\newcommand\hcall{\textsc{hc-all}}
\newcommand\shx{\textsc{sh-x}}
\newcommand\nosa{\textsc{-sa}}
\newcommand\noff{\textsc{-ff}}

\newcommand\blankfootnote[1]{%
  \let\thefootnote\relax\footnotetext{#1}%
  \let\thefootnote\svthefootnote%
}

\usepackage[shortlabels]{enumitem}
\setitemize{itemsep=2pt,topsep=2pt,parsep=0pt,partopsep=5pt}

\usepackage{url}

\title{Hard-Coded Gaussian Attention for Neural Machine Translation}

\author{Weiqiu You$^*$, Simeng Sun$^*$, Mohit Iyyer \\
  College of Information and Computer Sciences \\
  University of Massachusetts Amherst \\
  \texttt{\{wyou,simengsun,miyyer\}@cs.umass.edu} \\}

\date{}

\begin{document}
\maketitle

\begin{abstract}

Recent work has questioned the importance of the Transformer's multi-headed attention for achieving high translation quality. We push further in this direction by developing a ``hard-coded'' attention variant without \textit{any} learned parameters. Surprisingly, replacing all learned self-attention heads in the encoder and decoder with fixed, input-agnostic Gaussian distributions minimally impacts BLEU scores across four different language pairs. However, additionally hard-coding cross attention (which connects the decoder to the encoder) significantly lowers BLEU, suggesting that it is more important than self-attention. Much of this  BLEU drop can be recovered by adding just a \textit{single} learned cross attention head to an otherwise hard-coded Transformer. Taken as a whole, our results offer insight into which components of the Transformer are actually important, which we hope will guide future work into the development of simpler and more efficient attention-based models. 

\blankfootnote{* Authors contributed equally.}


\end{abstract}
\section{Introduction}\label{sec:intro}

The Transformer~\citep{NIPS2017_7181} has become the architecture of choice for neural machine translation. Instead of using recurrence to contextualize source and target token representations, Transformers rely on multi-headed attention mechanisms (MHA), which speed up training by enabling parallelization across timesteps. Recent work has called into question how much MHA contributes to translation quality: for example, a significant fraction of attention heads in a pretrained Transformer can be pruned without appreciable loss in BLEU~\citep{voita-etal-2019-analyzing,NIPS2019_9551}, and self-attention can be replaced by less expensive modules such as convolutions~\citep{yang-etal-2018-modeling,wu2018pay}. 

\begin{figure}[ht!]
    \centering
    \includegraphics[width=\linewidth]{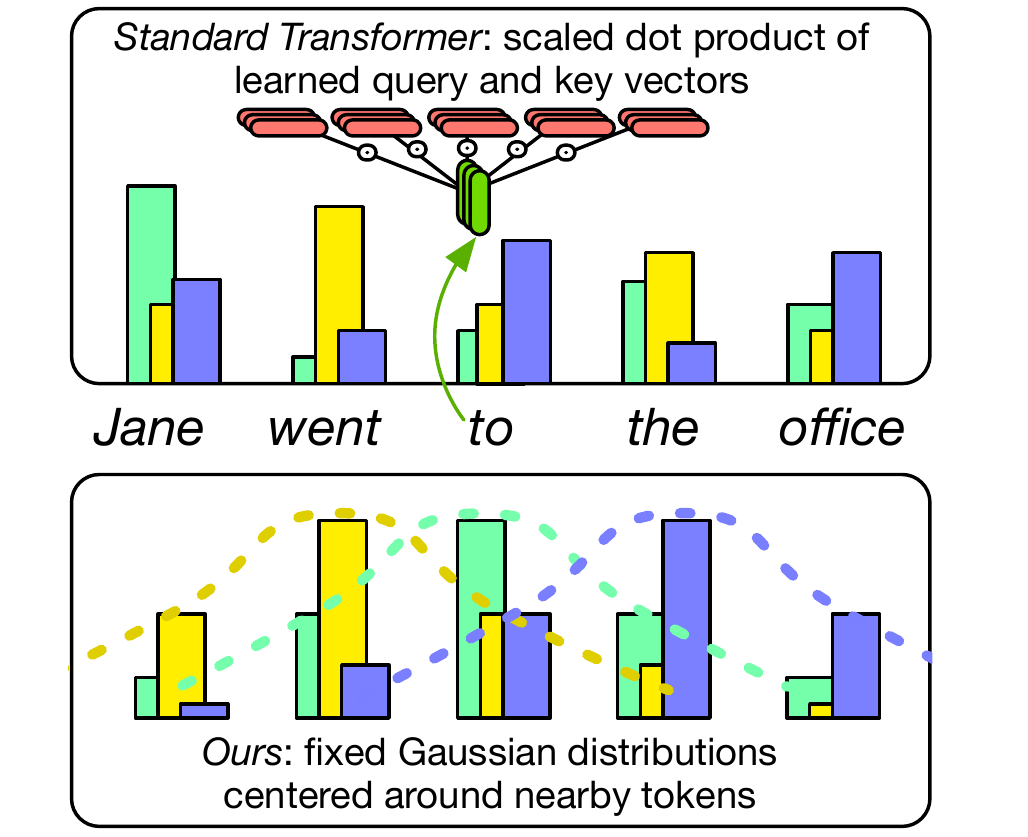}
    \caption{Three heads of learned self-attention (top) as well as our hard-coded attention (bottom) given the query word ``to''. In our variant, each attention head is a Gaussian distribution centered around a different token within a local window.}
    \label{fig:hcattn}
\end{figure}

In this paper, we take this direction to an extreme by developing a variant of MHA without \textit{any} learned parameters (Section~\ref{sec:model}). Concretely, we replace each attention head with a ``hard-coded'' version, which is simply a standard normal distribution centered around a particular position in the sequence (Figure~\ref{fig:hcattn}).\footnote{In Figure~\ref{fig:hcattn}, the hard-coded head distribution centered on the word ``to'' (shown in green) is  $[0.054, 0.24, 0.40, 0.24, 0.054]$.} When we replace all encoder and decoder self-attention mechanisms with our hard-coded variant, we achieve almost identical BLEU scores to the baseline Transformer for four different language pairs 
(Section~\ref{sec:experiments}).\footnote{Our code is available at \url{https://github.com/fallcat/stupidNMT}}

\begin{figure*}[ht!]
    \centering
    \resizebox{1\textwidth}{!}{%
    \includegraphics[width=\textwidth]{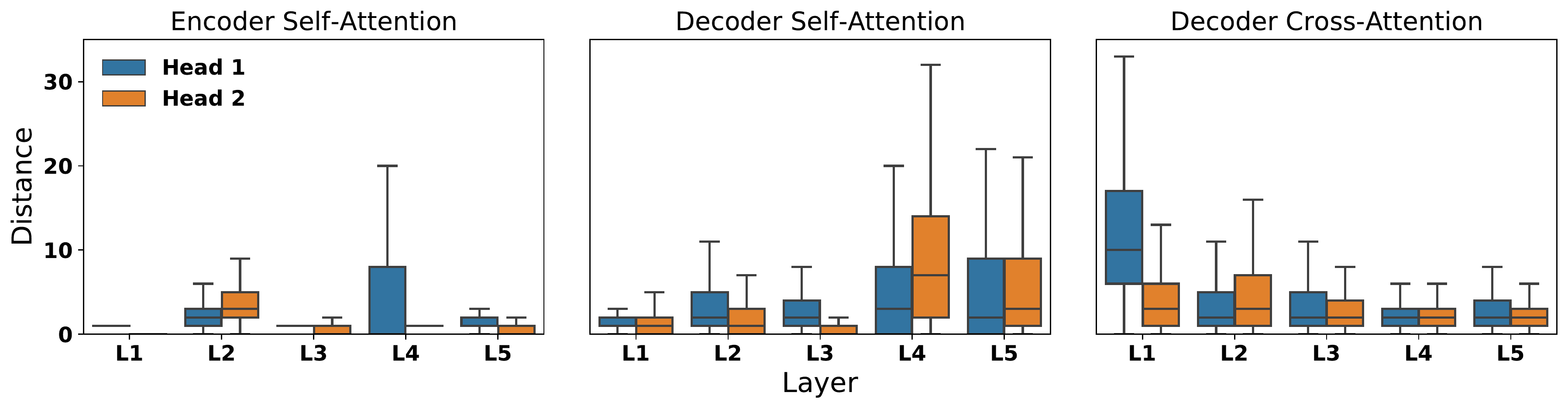}
    }
    \caption{Most learned attention heads for a Transformer trained on IWSLT16 En-De focus on a local window around the query position. The x-axis plots each head of each layer, while the y-axis refers to the distance between the query position and the argmax of the attention head distribution (averaged across the entire dataset).}
    \label{fig:attn-stats}
\end{figure*}

These experiments maintain fully learned MHA \textit{cross attention}, which allows the decoder to condition its token representations on the encoder's outputs. We next attempt to additionally replace cross attention with a hard-coded version, which results in substantial drops of 5-10 BLEU. Motivated to find the minimal number of learned attention parameters needed to make up this deficit, we explore configurations with only \textit{one} learned cross attention head in total, which performs just slightly worse (1-3 BLEU) than the baseline.

By replacing MHA with hard-coded attention, we improve memory efficiency (26.4\% more tokens per batch) and decoding speed (30.2\% increase in sentences decoded per second)  without significantly lowering BLEU, although these efficiency improvements are capped by other more computationally-expensive components of the model (Section~\ref{sec:efficiency}). We also perform analysis experiments (Section~\ref{subsec:erroranalysis}) on linguistic properties (e.g., long-distance subject-verb agreement) that MHA is able to better model than hard-coded attention. Finally, we develop further variants of hard-coded attention in Section~\ref{subsec:modifications}, including a version without any attention weights at all.   

Our hard-coded Transformer configurations have intuitively severe limitations: attention in a particular layer is highly concentrated on a local window in which fixed weights determine a token's importance. Nevertheless, the strong performance of these limited models indicates that the flexibility enabled by fully-learned MHA is not as crucial as commonly believed: perhaps  attention is not \textit{all} you need. We hope our work will spur further development of simpler, more efficient models for neural machine translation.

\section{Background}\label{sec:background}

In this section, we first briefly review the Transformer architecture of~\citet{NIPS2017_7181} with a focus on its multi-headed attention. Then, we provide an analysis of the learned attention head distributions of a trained Transformer model, which motivates the ideas discussed afterwards. 

\subsection{Multi-headed Transformer attention}
The Transformer is an encoder-decoder model formed by stacking layers of attention blocks. Each encoder block contains a self-attention layer followed by layer normalization, a residual connection, and a feed-forward layer. Decoder blocks are identical to those of the encoder except they also include a cross attention layer, which connects the encoder's representations to the decoder. 

To compute a single head of self-attention given a sequence of token representations $\bvec{t}_{1\dots n}$, we first project these representations to queries $\bvec{q}_{1\dots n}$, keys $\bvec{k}_{1\dots n}$, and values $\bvec{v}_{1\dots n}$ using three different linear projections. Then, to compute the self-attention distribution at a particular position $i$ in the sequence, we take the scaled dot product between the query vector $\bvec{q}_i$ and all of the key vectors (represented by matrix \bmat{K}). We then use this distribution to compute a weighted average of the values (\bmat{V}):
\begin{equation}
    \text{Attn}(\bvec{q}_i, \bmat{K}, \bmat{V}) = \text{softmax}(\frac{\bvec{q}_i\bmat{K}^\top}{\sqrt{d_k}})\bmat{V}
    \label{eq:attn_weights}
\end{equation}
\noindent where $d_k$ is the dimensionality of the key vector. 

For MHA, we use different projection matrices to obtain the query, key, and value representations for each head. The key difference between self-attention and cross attention is that the queries and keys come from different sources: specifically, the keys are computed by passing the encoder's final layer token representations through a linear projection. To summarize, MHA is used in three different components of the Transformer: encoder self-attention, decoder self-attention, and cross attention.  


\subsection{Learned heads mostly focus on local windows}
The intuition behind MHA is that each head can focus on a different type of information (e.g., syntactic or semantic patterns). While some heads have been shown to possess interpretable patterns \cite{voita-etal-2019-analyzing, correia-etal-2019-adaptively}, other work has cautioned against using attention patterns to explain a model's behavior \citep{jain-wallace-2019-attention}. In our analysis, we specifically examine the behavior of a head with respect to the current query token's position in the sequence. We train a baseline Transformer model (five layers, two heads per layer) on the IWSLT 2016 En$\rightarrow$De dataset, and compute aggregated statistics on its learned heads.


Figure~\ref{fig:attn-stats} shows that outside of a few layers, most of the model's heads focus their attention (i.e., the argmax of the attention distribution) on a local neighborhood around the current sequence position. For example, both self-attention heads in the first layer of the encoder tend to focus on just a one to two token window around the current position.  The decoder self-attention and cross attention heads show higher variability, but most of their heads are still on average focused on local information.  These results beg the question of whether replacing self-attention with ``hard-coded'' patterns that focus on local windows will significantly affect translation quality.

\section{Hard-coded Gaussian attention}\label{sec:model}
While learned attention enables model flexibility (e.g., a head can ``look'' far away from the current position if it needs to), it is unclear from the above analysis how crucial this flexibility is. To examine this question, we replace the attention distribution computation in Equation~\ref{eq:attn_weights} (i.e., scaled dot product of queries and keys)  with a fixed Gaussian distribution.\footnote{~\citet{yang-etal-2018-modeling} implement a similar idea, except the mean and standard deviation of their Gaussians are learned with separate neural modules.} In doing so, we remove \textit{all} learned parameters from the attention computation: the mean of the Gaussian is determined by the position $i$ of the current query token, and the standard deviation is always set to 1.\footnote{Preliminary experiments with other standard deviation values did not yield significant differences, so we do not vary the standard deviation for any experiments in this paper.} As Transformers contain both self-attention and cross attention, the rest of this section details how we replace both of these components with simplified versions. We will refer to experimental results on the relatively small IWSLT16 English-German dataset throughout this section to contextualize the impact of the various design decisions we describe. Section~\ref{sec:experiments} contains a more fleshed out experimental section with many more datasets and language pairs.

\subsection{Hard-coded self-attention}
\label{subsec:hcsa}
In self-attention, the queries and keys are derived from the same token representations and as such have the same length $n$. The baseline Transformer (\base) computes the self-attention distribution at position $i$ by taking the dot product between the query representation $\bvec{q}_i$ and all of the key vectors $\bvec{k}_{1 \dots n}$. We instead use a fixed Gaussian distribution centered around position $i-1$ (token to the left), $i$ (the query token), or $i+1$ (token to the right). More formally, we replace Equation~\ref{eq:attn_weights} with
\begin{equation}
    \text{Attn}(i, \bmat{V}) =  \mathcal{N}(f(i), \sigma^2)\bmat{V}.
\end{equation}
The mean of the Gaussian $f(i)$ and its standard deviation $\sigma^2$ are both hyperparameters; for all of our experiments, we set $\sigma$ to 1 and $f(i)$ to either $i-1$, $i$  or $i+1$, depending on the head configuration.\footnote{The Gaussian distribution is cut off on the borders of the sentence and is not renormalized to sum to one.} Note that this definition is completely agnostic to the input representation: the distributions remain the same regardless of what sentence is fed in or what layer we are computing the attention at. Additionally, our formulation removes the query and key projections from the attention computation; the Gaussians are used to compute a weighted average of the value vectors.\footnote{Preliminary models that additionally remove the value projections performed slightly worse when we hard-coded cross attention, so we omit them from the paper.}

Instead of learning different query and key projection matrices to define different heads, we simply design head distributions with different means. Figure~\ref{fig:hcattn} shows an example of our hard-coded self-attention for a simple sentence. We iterate over different configurations of distribution means $f(i)$ on the IWSLT16 En-De dataset, while keeping the cross attention learned.\footnote{See Appendix for a table describing the effects of varying $f(i)$ on IWSLT16 En-De BLEU score. We find in general that hard-coded heads within each layer should focus on different tokens within the local window for optimal performance.}
Our best validation result with hard-coded self-attention (\hcsa) replaces encoder self-attention with distributions centered around $i-1$ and $i+1$ and decoder self-attention with distributions centered around $i-1$ and $i$. This model achieves slightly \textit{higher} BLEU than the baseline Transformer (\textbf{30.3} vs \textbf{30.0} BLEU).

\subsection{Alternatives to cross attention}
We turn next to cross attention, which on its face seems more difficult to replace with hard-coded distributions. Unlike self-attention, the queries and keys in cross attention are not derived from the same token representations; rather, the queries come from the decoder while the keys come from the encoder. Since the number of queries can now be different from the number of keys, setting the distribution means by position is less trivial than it is for self-attention. Here, we describe two methods to simplify cross attention, starting with a fully hard-coded approach and moving onto a minimal learned configuration. 

\paragraph{Hard-coded cross attention:} We begin with a simple solution to the problem of queries and keys having variable lengths. Given a training dataset, we compute the length ratio $\gamma$ by dividing the average source sentence length by the average target sentence length. Then, to define a hard-coded cross attention distribution for target position $i$, we center the Gaussian on positions $\floor{\gamma i - 1}$, $\floor{\gamma i}$, and $\floor{\gamma i + 1}$ of the source sentence. When we implement this version of hard-coded cross attention and also hard-code the encoder and decoder self-attention as described previously (\hcall), our BLEU score on IWSLT16 En-De drops from \textbf{30.3} to \textbf{21.1}. Clearly, cross attention is more important for maintaining translation quality than self-attention.~\citet{NIPS2019_9551} notice a similar phenomenon when pruning heads from a pretrained Transformer: removing certain cross attention heads can substantially lower BLEU.

\paragraph{Learning a single cross attention head:} 
Prior to the advent of the Transformer, many neural machine translation architectures relied on just a single cross attention ``head''~\citep{DBLP:journals/corr/BahdanauCB14}. The Transformer has many heads at many layers, but how many of these are actually necessary? Here, we depart from the parameter-free approach by instead removing cross attention at all but the final layer of the decoder, where we include only a single learned head (\shx). Note that this is the only learned head in the entire model, as both the encoder and decoder self-attention is hard-coded. On IWSLT16 En-De, our BLEU score improves from \textbf{21.1} to \textbf{28.1}, less than 2 BLEU under the \base\ Transformer.

\section{Large-scale Experiments} \label{sec:experiments}

\begin{table}[t!]
    \centering
    \scalebox{0.75}{
    \begin{tabular}{l*{4}{c}}
    \toprule
         & Train & Test & Len SRC & Len TGT\\ \midrule
      IWSLT16 En-De & 196,884 & 993 & 28.5 & 29.6  \\
      IWSLT17 En-Ja & 223,108 & 1,452 & 22.9 & 16.0 \\
        WMT16 En-Ro & 612,422 & 1,999 & 27.4 & 28.3 \\
        WMT14 En-De & 4,500,966 & 3,003 & 28.5 & 29.6\\
        WMT14 En-Fr & 10,493,816 & 3,003 & 26.0 & 28.8\\
    \bottomrule
    \end{tabular}
    }
    \caption{Statistics of the datasets used. The last two columns show the average number of tokens for source and target sentences, respectively. }
    \label{tab:dataset}
\end{table}

The previous section developed hard-coded configurations and  presented results on the relatively small IWSLT16 En-De dataset. Here, we expand our experiments to include a variety of different datasets, language pairs, and model sizes. For all hard-coded head configurations, we use the optimal IWSLT16 En-De setting detailed in Section~\ref{subsec:hcsa} and perform no additional tuning on the other datasets. This configuration nevertheless proves robust, as we observe similar trends with our hard-coded Transformers across all of datasets.\footnote{Code and scripts to reproduce our experimental results to be released after blind review.}

\subsection{Datasets}
We experiment with four language pairs, English$\leftrightarrow$\{German, Romanian, French, Japanese\} to show the consistency of our proposed attention variants. For the En-De pair, we use both the small IWSLT 2016\footnote{We report BLEU on the IWSLT16 En-De dev set following previous work~\citep{gu2018nonautoregressive,lee-etal-2018-deterministic,  akoury-etal-2019-syntactically}. For other datasets, we report test BLEU.} and the larger WMT 2014 datasets. For all datasets except WMT14 En$\rightarrow$De and WMT14 En$\rightarrow$Fr,\footnote{As the full WMT14 En$\rightarrow$Fr is too large for us to feasibly train on, we instead follow~\citet{akoury-etal-2019-syntactically} and train on just the Europarl / Common Crawl subset, while evaluating
using the full dev/test sets.} we run experiments in both directions. For English-Japanese, we train and evaluate on IWSLT 2017 En$\leftrightarrow$Ja TED talk dataset. More dataset statistics are shown in Table~\ref{tab:dataset}.

\subsection{Architectures}
Our \base\ model is the original Transformer from~\citet{NIPS2017_7181}, reimplemented in PyTorch \cite{NIPS2019_9015} by \citet{akoury-etal-2019-syntactically}.\footnote{\url{https://github.com/dojoteef/synst}} To implement hard-coded attention, we only modify the attention functions in this codebase and keep everything else the same. For the two small IWSLT datasets, we follow prior work by using a small Transformer architecture with embedding size 288, hidden size 507, four heads,\footnote{For hard-coded configurations, we duplicate heads to fit this architecture (e.g., we have two heads per layer in the encoder with means of $i+1$ and $i-1$).} five layers, and a learning rate 3e-4 with a linear scheduler. For the larger datasets, we use the standard Tranformer base model, with embedding size 512, hidden size 2048, eight heads, six layers, and a warmup scheduler with 4,000 warmup steps. For all experiments, we report BLEU scores using SacreBLEU \citep{post-2018-call} to be able to compare with other work.\footnote{SacreBLEU signature: BLEU+case.mixed+lang.LANG
+numrefs.1+smooth.exp+test.TEST+tok.intl+version.1.2.11,
with LANG $\in$ \{en-de, de-en, en-fr\} and TEST $\in$ \{wmt14/full,
iwslt2017/tst2013\}. For WMT16 En-Ro and IWSLT17 En-Ja, we follow previous work for preprocessing \citep{sennrich-etal-2016-edinburgh}, encoding the latter with a 32K sentencepiece vocabulary (\url{https://github.com/google/sentencepiece}) and measuring the de-tokenized BLEU with SacreBLEU.} 

\begin{table}[t!]
    \centering
    \scalebox{0.85}{
    \begin{tabular}{lcccc}
    \toprule
         & \base & \hcsa & \hcall & \shx \\ \midrule
        IWSLT16 En-De & 30.0 & 30.3 & 21.1 & 28.2 \\
        IWSLT16 De-En & 34.4 & 34.8 & 25.7 & 33.3 \\
        IWSLT17 En-Ja & 20.9 & 20.7 & 10.6 & 18.5 \\
        IWSLT17 Ja-En & 11.6 & 10.9 & 6.1 & 10.1 \\
        WMT16 En-Ro & 33.0 & 32.9 & 25.5 & 30.4 \\
        WMT16 Ro-En & 33.1 & 32.8 & 26.2 & 31.7 \\ \midrule
        WMT14 En-De & 26.8 & 26.3 & 21.7 & 23.5 \\
        WMT14 En-Fr & 40.3 & 39.1 & 35.6 & 37.1 \\
    \bottomrule
    \end{tabular}
    }
    \caption{Comparison of the discussed Transformer variants on six smaller datasets (top)\footnote{The low performance on En-Ja may be due to the type of tokenizer used and the number of vocabulary.} and two larger datasets (bottom). Hard-coded self-attention (\hcsa) achieves almost identical BLEU scores to \base\ across all datasets, while a model with only one cross attention head (\shx) performs slightly worse. }
    \label{tab:hard-coded-self-attn}
\end{table}

\subsection{Summary of results} \label{section:hard-coded-attention}
Broadly, the trends we observed on IWSLT16 En-De in the previous section are consistent for all of the datasets and language pairs. Our findings are summarized as follows:
\begin{itemize}
    \item A Transformer with hard-coded self-attention in the encoder and decoder and learned cross attention (\hcsa) achieves almost equal BLEU scores to the \base\ Transformer.
    \item Hard-coding both cross attention and self-attention (\hcall) considerably drops BLEU compared to \base, suggesting cross attention is more important for translation quality.
    \item A configuration with hard-coded self-attention and a single learned cross attention head in the final decoder layer (\shx) consistently performs 1-3 BLEU worse than \base.
\end{itemize}

These results motivate a number of interesting analysis experiments (e.g., what kinds of phenomena is MHA better at handling than hard-coded attention), which we describe in Section~\ref{sec:analysis}. The strong performance of our highly-simplified models also suggests that we may be able to obtain memory or decoding speed improvements, which we investigate in the next section.

\section{Bigger Batches \& Decoding Speedups}\label{sec:efficiency}
We have thus far motivated our work as an exploration of which components of the Transformer are necessary to obtain high translation quality. Our results demonstrate that encoder and decoder self-attention can be replaced with hard-coded attention distributions without loss in BLEU, and that MHA brings minor improvements over single-headed cross attention. In this section, we measure efficiency improvements in terms of batch size increases and decoding speedup.

\paragraph{Experimental setup:} We run experiments on WMT16 En-Ro with the larger architecture to support our conclusions.\footnote{Experiments with the smaller IWSLT16 En-De model are described in the Appendix.} For each model variant discussed below, we present its memory efficiency as the maximum number of tokens per batch allowed during training on a single GeForce RTX 2080 Ti. Additionally, we provide inference speed as the number of sentences per second each model can decode on a 2080 Ti, reporting the average of five runs with a batch size of 256.

\paragraph{Hard-coding self-attention yields small efficiency gains:}
Table \ref{tab:mem-time-profile} summarizes our profiling experiments. Hard-coding self-attention and preserving learned cross attention allows us to fit 17\% more tokens into a single batch, while also providing a 6\% decoding speedup compared to \base\ on the larger architecture used for WMT16 En-Ro. 
The improvements in both speed and memory usage are admittedly limited, which motivates us to measure the maximum efficiency gain if we only modify self-attention (i.e., preserving learned cross attention). We run a set of upper bound experiments where we entirely remove self-attention in the encoder and decoder. The resulting encoder thus just becomes a stack of feed-forward layers on top of the initial subword embeddings. Somewhat surprisingly, the resulting model still achieves a fairly decent BLEU of \textbf{27.0} compared to the \base\ model's \textbf{33.0}. 
As for the efficiency gains, we can fit 27\% more tokens into a single batch, and decoding speed improves by 12.3\% over \base. This relatively low upper bound for \hcsa\ shows that simply hard-coding self-attention does not guarantee significant speedup. Previous work that simplifies attention~\cite{wu2018pay,NIPS2019_9551}  also report efficiency improvements of similar low magnitudes.

\paragraph{Single-headed cross attention speeds up decoding:}
Despite removing learned self-attention from both the encoder and decoder, we did not observe huge efficiency or speed gains. However,  reducing the source attention to just a single head results in more significant improvements. By only keeping single-headed cross attention in the last layer, we are able to achieve 30.2\% speed up and fit in 26.4\% more tokens to the memory compared to \base\ . Compared to \hcsa, \shx\ obtains a 22.9\% speedup and 8.0\% bigger batch size.


From our profiling experiments, most of the speed and memory considerations of the Transformer are associated with the large feed-forward layers that we do not modify in any of our experiments, which caps the efficiency gains from modifying the attention implementation. While we did not show huge efficiency improvements on modern GPUs, it remains possible that (1) a more tailored implementation could leverage the model simplifications we have made, and (2)  that these differences are larger on other hardware (e.g., CPUs). We leave these questions for future work.

\begin{table}[t!]
    \centering
    \begin{tabular}{cccc} 
        \toprule
          Model & BLEU & sent/sec & tokens/batch \\ \midrule
         \base & 33.0 & 26.8 & 9.2K \\
         \hcsa & 32.9 & 28.4 & 10.8K \\
         \shx & 30.3 & 34.9 & 11.7K \\
         \midrule
        \base /\nosa & 27.0 & 30.1 & 11.8K \\
        \shx /\nosa & 15.0 & 37.6 & 13.3K \\
        \bottomrule
    \end{tabular}
    \caption{{Decoding speedup (in terms of sentences per second) and memory improvements (max tokens per batch) on WMT16 En-Ro for a variety of models. The last two rows refer to \base\ and \shx\ configurations whose self-attention is completely removed. }}
    \label{tab:mem-time-profile}
\end{table}


\section{Analysis}\label{sec:analysis}
Taken as a whole, our experimental results suggest that many of the components in the Transformer can be replaced by highly-simplified versions without adversely affecting translation quality. In this section, we explain how hard-coded self-attention does not degrade translation quality (Section~\ref{subsec:feedforward-analysis}),
perform a detailed analysis of the behavior of our various models by comparing the types of errors made by learned versus hard-coded attention (Section~\ref{subsec:erroranalysis}), and also examine different  attention configurations that naturally follow from our experiments (Section~\ref{subsec:modifications}).

\begin{figure}[t!]
    \centering
    \includegraphics[width=0.45\textwidth]{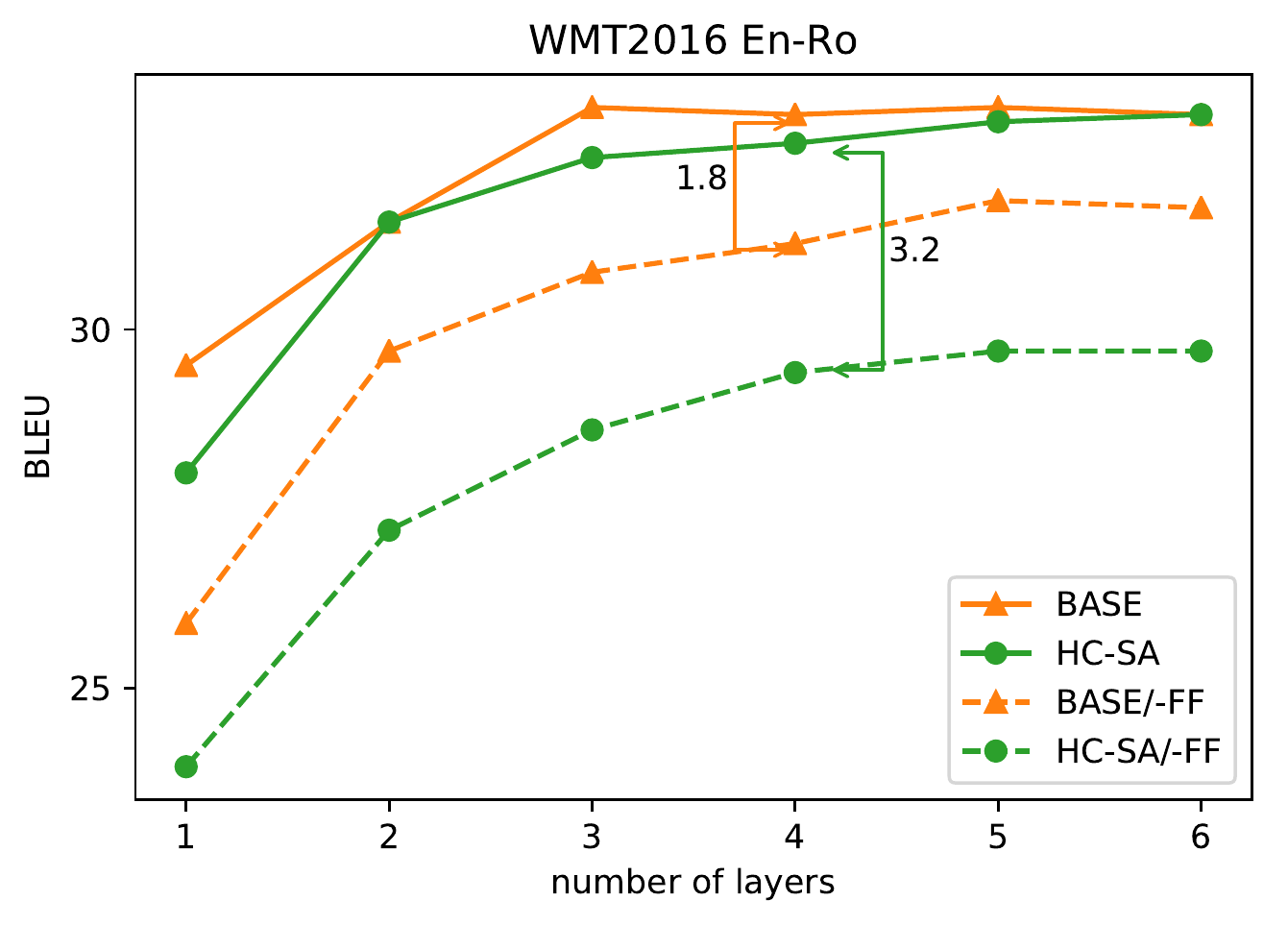}
    \caption{BLEU performance on WMT16 En-Ro before and after removing all feed-forward layers from the models. \base\ and \hcsa\ achieve almost identical BLEU scores, but \hcsa\ relies more on the feed-forward layers than the vanilla Transformer. As shown on the plot, with a four layer encoder and decoder, the BLEU gap between \base\noff\ and \base\ is 1.8, while the gap between \hcsa\ and \hcsa\noff\ is 3.2.}
    \label{fig:enro-ffn}
\end{figure}

\subsection{Why does hard-coded self-attention work so well?} \label{subsec:feedforward-analysis}
Given the good performance of \hcsa\ on multiple datasets, it is natural to ask why hard-coding self-attention does not deteriorate translation quality. We conjecture that feed-forward (FF) layers play a more important role in \hcsa\ than in \base\ by compensating for the loss of learned dynamic self-attention. To test this hypothesis, we conduct an analysis experiment in which we train four model configurations while varying the number of layers: \base, \base\ without feed-forward layers (\base/\noff), \hcsa\ and \hcsa without feed-forward layers (\hcsa/\noff). As shown in Figure~\ref{fig:enro-ffn}, \base\ and \hcsa\ have similar performance and both \noff\ models have consistently lower BLEU scores. However, \hcsa\ without FF layers performs much worse compared to its \base\ counterpart. This result confirms our hypothesis that FF layers are more important in \hcsa\ and capable of recovering the potential performance degradation brought by hard-coded self-attention. Taking a step back to hard-coding cross attention, the failure of hard-coding cross attention might be because the feed-forward layers of the decoder are not powerful enough to compensate for modeling both hard-coded decoder self-attention and  cross attention.

\subsection{Error analysis of hard-coded models}
\label{subsec:erroranalysis}

\begin{figure}[t!]
    \centering
    \includegraphics[width=0.45\textwidth]{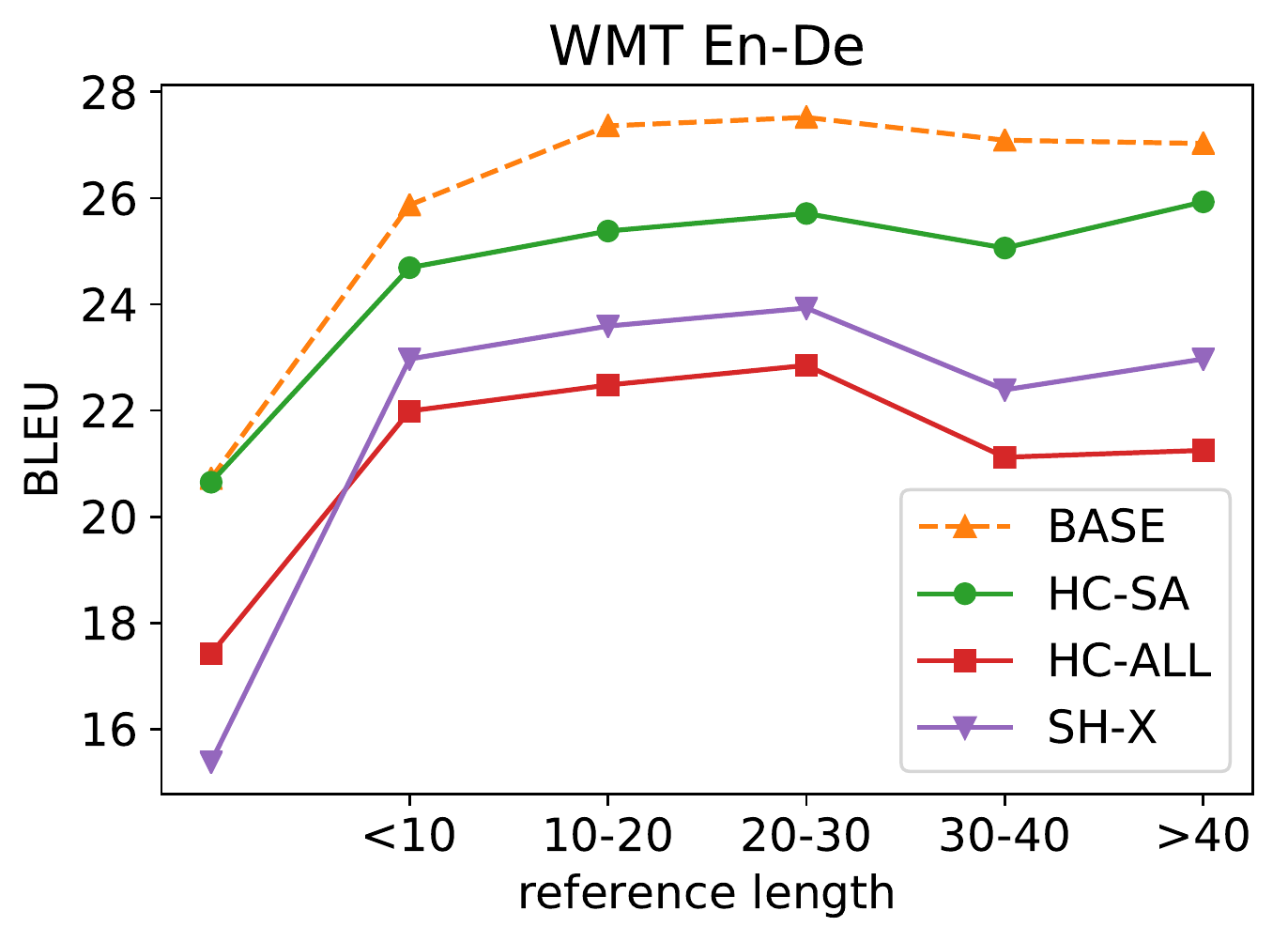}
    \caption{BLEU difference vs. \base\ as a function of reference length on the WMT14 En-De test set. When cross attention is hard-coded (\hcall), the BLEU gap worsens as reference length increases.}
    \label{fig:bleu-vs-len}
\end{figure}

\paragraph{Is learned attention more important for longer sentences?} \label{section:bleu-vs-len}
Since hard-coded attention is much less flexible than learned attention and can struggle to encode global information, we are curious to see if its performance declines as a function of sentence length. To measure this, we categorize the WMT14 En-De test set into five bins by reference length and plot the decrease in BLEU between \base\ and our hard-coded configurations for each bin. Somewhat surprisingly, Figure~\ref{fig:bleu-vs-len} shows that the BLEU gap between \base\ and \hcsa\ seems to be roughly constant across all bins.\footnote{We note that gradients will flow across long distances if the number of layers is large enough, since the effective window size increases with multiple layers~\cite{DBLP:conf/ssw/OordDZSVGKSK16, DBLP:journals/corr/KalchbrennerESO16}.} However, the fully hard-coded \hcall\ model clearly deteriorates as reference length increases.

\paragraph{Does hard-coding attention produce any systematic linguistic errors?}

\begin{table}[ht!]
    \centering
    \scalebox{0.85}{
    \begin{tabular}{lccc}
    \toprule
        Error type & \base & \hcsa & \hcall\\\midrule
        np-agreement & \textbf{54.2} & 53.5 & 53.5\\
        subj-verb-agreement & \textbf{87.5} & 85.8 & 82.5\\
        subj-adequacy & \textbf{87.3} & 85.0 & 80.3\\
        polarity-particle-nicht-del & \textbf{94.0} & 91.4 & 83.2\\
        polarity-particle-kein-del & \textbf{91.4} & 88.3 & 79.9\\
        polarity-affix-del & \textbf{91.6} & 90.8 & 83.1\\
        polarity-particle-nicht-ins & \textbf{92.6} & 92.5 & 89.8\\
        polarity-particle-kein-ins & 94.8 & 96.7 & \textbf{98.7}\\
        polarity-affix-ins & \textbf{91.9} & 90.6 & 84.3\\
        auxiliary & \textbf{89.1} & 87.5 & 85.6\\
        verb-particle & \textbf{74.7} & 72.7 & 70.2\\
        compound & 88.1 & \textbf{89.5} & 80.5\\
        transliteration & 97.6 & \textbf{97.9} & 93.4\\
        \bottomrule
    \end{tabular}
    }
    \caption{Accuracy for each error type in the  LingEval97 contrastive set. Hard-coding self-attention results in slightly lower accuracy for most error types, while more significant degradation is observed when hard-coding self and cross attention. We refer readers to ~\citet{sennrich-2017-grammatical} for descriptions of each error type.}
    \label{tab:lingeval-err}
\end{table} 

\begin{figure}[t!]
    \centering
    \includegraphics[width=0.44\textwidth]{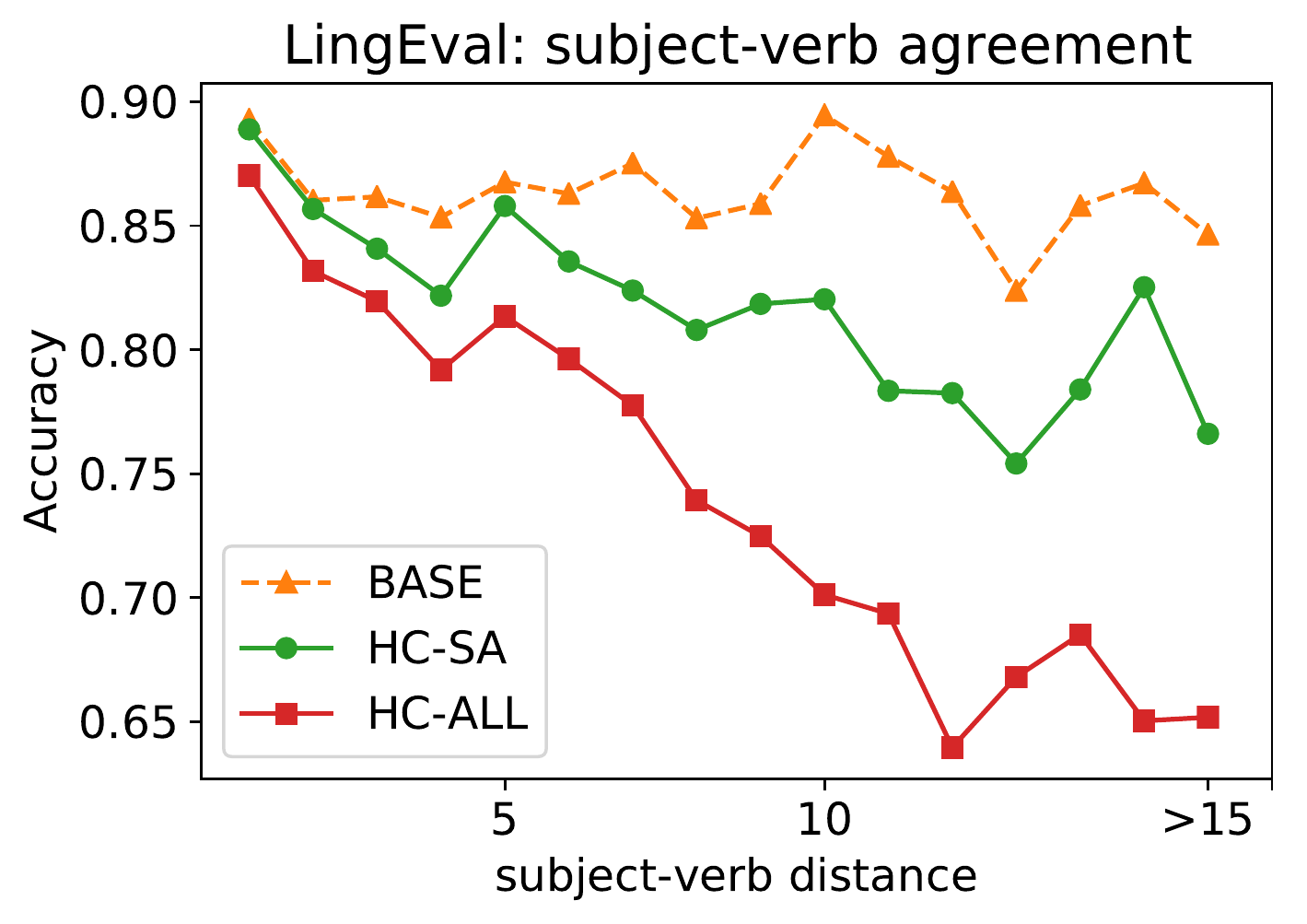}
    \caption{Hard-coded models become increasingly worse than 
    \base\ at subject-verb agreement as the dependency grows longer. }
    \label{fig:lingeval-subj-verb}
\end{figure}

For a more fine-grained analysis, we run experiments on LingEval97~\citep{sennrich-2017-grammatical}, an English$\rightarrow$German dataset consisting of contrastive translation pairs. This dataset measures targeted errors on thirteen different linguistic phenomena such as agreement and adequacy.
\base\ and \hcsa\ perform\footnote{Accuracy is computed by counting how many references have lower token-level cross entropy loss than their contrastive counterparts.} very similarly across all error types (Table \ref{tab:lingeval-err}), which is perhaps unsurprising given that their BLEU scores are almost identical. Interestingly, the category with the highest decrease from \base\ for both \hcsa\ and \hcall\ is \textit{deleted negations};\footnote{Specifically, when \textit{ein} is replaced with negation \textit{kein}.} \hcall\ is  11\% less accurate (absolute) at detecting these substitutions than \base\ (94\% vs 83\%). On the other hand, both \hcsa\ and \hcall\ are actually better than \base\ at detecting \textit{inserted negations}, with \hcall\ achieving a robust 98.7\% accuracy. We leave further exploration of this phenomenon to future work. Finally, we observe that for the subject-verb agreement category, the discrepancy between \base\ and the hard-coded models increases as the distance between subject-verb increases (Figure \ref{fig:lingeval-subj-verb}). This result confirms that self-attention is important for modeling some long-distance phenomena, and that cross attention may be even more crucial.

\begin{figure}[t!]
    \centering
    \includegraphics[width=0.44\textwidth]{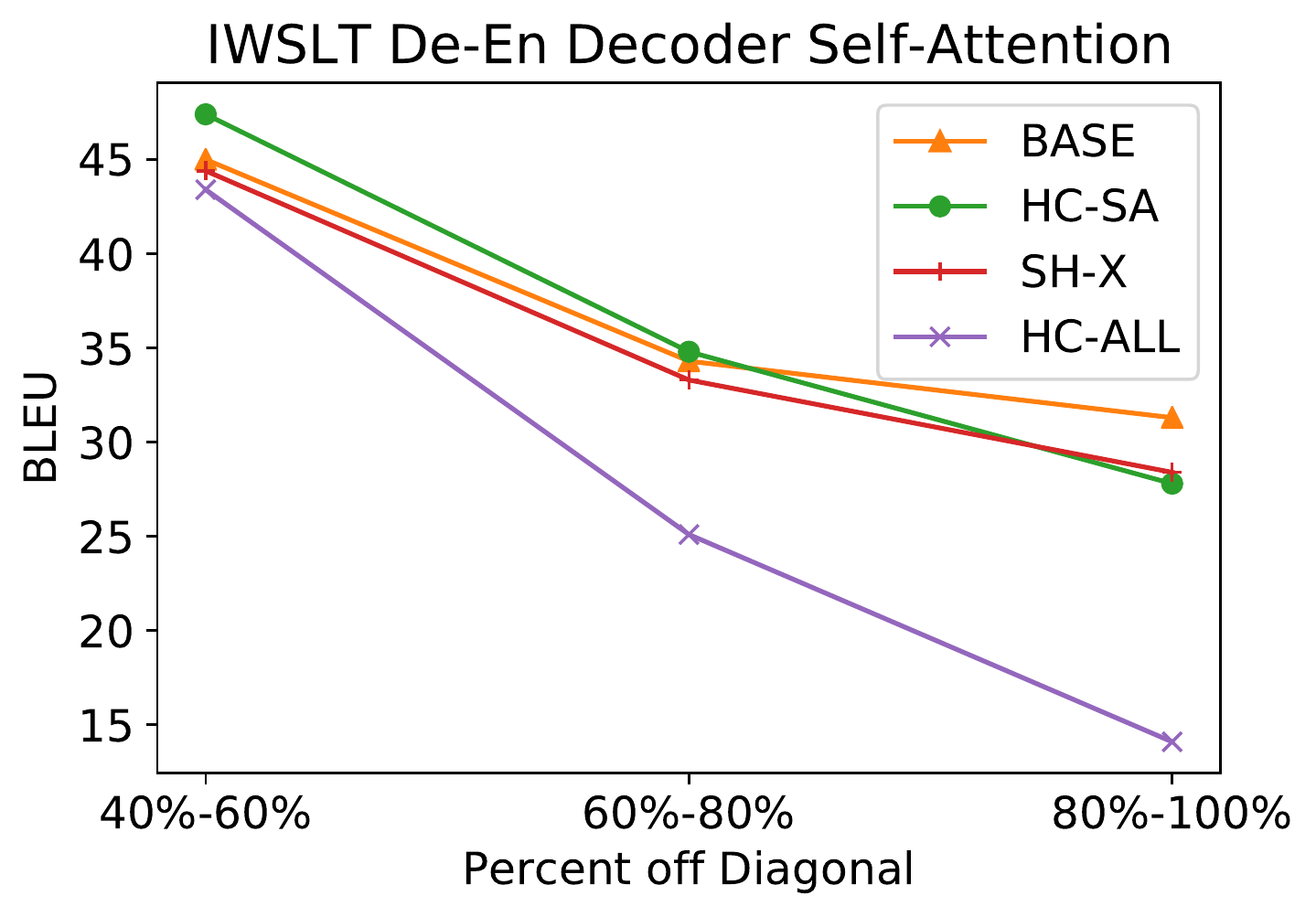}
    \caption{Hard-coded attention performs better for sentences with low off-diagonality (i.e., sentences for which the \base\ model's learned attention focuses close to the query position for most of their tokens).}
    \label{fig:off-diagonal}
\end{figure}

\paragraph{Do hard-coded models struggle when learned self-attention focuses on non-local information?}
Since hard-coded models concentrate most of the attention probability mass on local tokens, they might underperform on sentences for which the learned heads of the \base\ model focus on tokens far from the current query position. We define a token to be ``off-diagonal'' when the maximum probability of that token's attention is at least two steps away from query position. A sentence's ``off-diagonality'' is then the proportion of off-diagonal tokens within the sentence. We bin the sentences in IWSLT En-De development set by their  off-diagonality and analyze the translation quality of our models on these different bins. Figure~\ref{fig:off-diagonal} shows that for decoder self attention, the BLEU gap between \hcall\ and \base\ increases as off-diagonality increases, while the gap between \base\ and \shx\ remains relatively constant across all bins. \hcsa\ even outperforms \base\ for sentences with fewer off-diagonal tokens.

\subsection{Other hard-coded model configurations}
\label{subsec:modifications}

\begin{table}[t!]
    \centering
    \scalebox{0.9}{
        \begin{tabular}{cccc}
        \toprule
              & Original & Conv (window=3) & Indexing \\ \hline
             En-De & 30.3 & 30.1 & 29.8\\
             En-Ro & 32.4 & 32.3 & 31.4\\
         \bottomrule
        \end{tabular}
    }
    \caption{Comparison of three implementations of \hcsa. Truncating the distribution to a three token span has little impact, while removing the weights altogether slightly lowers BLEU.}
    \label{tab:diff-impl-bleu}
\end{table}
\paragraph{Is it important for the Gaussian to span the entire sequence?}
One natural question about the hard-coded attention strategy described in Section~\ref{sec:model} is whether it is necessary to assign some probability to all tokens in the sequence. After all, the probabilities outside a local window become very marginal, so perhaps it is unnecessary to preserve them. We take inspiration from~\citet{wu2018pay}, who demonstrate that lightweight convolutions can replace self-attention in the Transformer without harming BLEU, by recasting our hard-coded attention as a convolution with a hard-coded 1-D kernel.
While this decision limits the Gaussian distribution to span over just tokens within a fixed window around the query token, it does not appreciably impact BLEU (second column of Table \ref{tab:diff-impl-bleu}). We set the window size to 3 in all experiments, so the kernel weights become $[0.242, 0.399, 0.242]$. 


\paragraph{Are any attention weights necessary at all?} 
The previous setting with constrained window size suggests another follow-up:
is it necessary to have any attention weights within this local window at all? A highly-efficient alternative is to have each head simply select a single value vector associated with a token in the window. Here, our implementation requires no explicit multiplication with a weight vector, as we can compute each head's representation by simply indexing into the value vectors. Mathematically, this is equivalent to convolving with a binary kernel (e.g., convolution with $[1,0,0]$ is equivalent to indexing the left token representation). The third column of Table \ref{tab:diff-impl-bleu} shows that this indexing approach results in less than 1 BLEU drop across two datasets, which offers an interesting avenue for future efficiency improvements.

\begin{figure}[t!]
    \centering
    \includegraphics[width=0.45\textwidth]{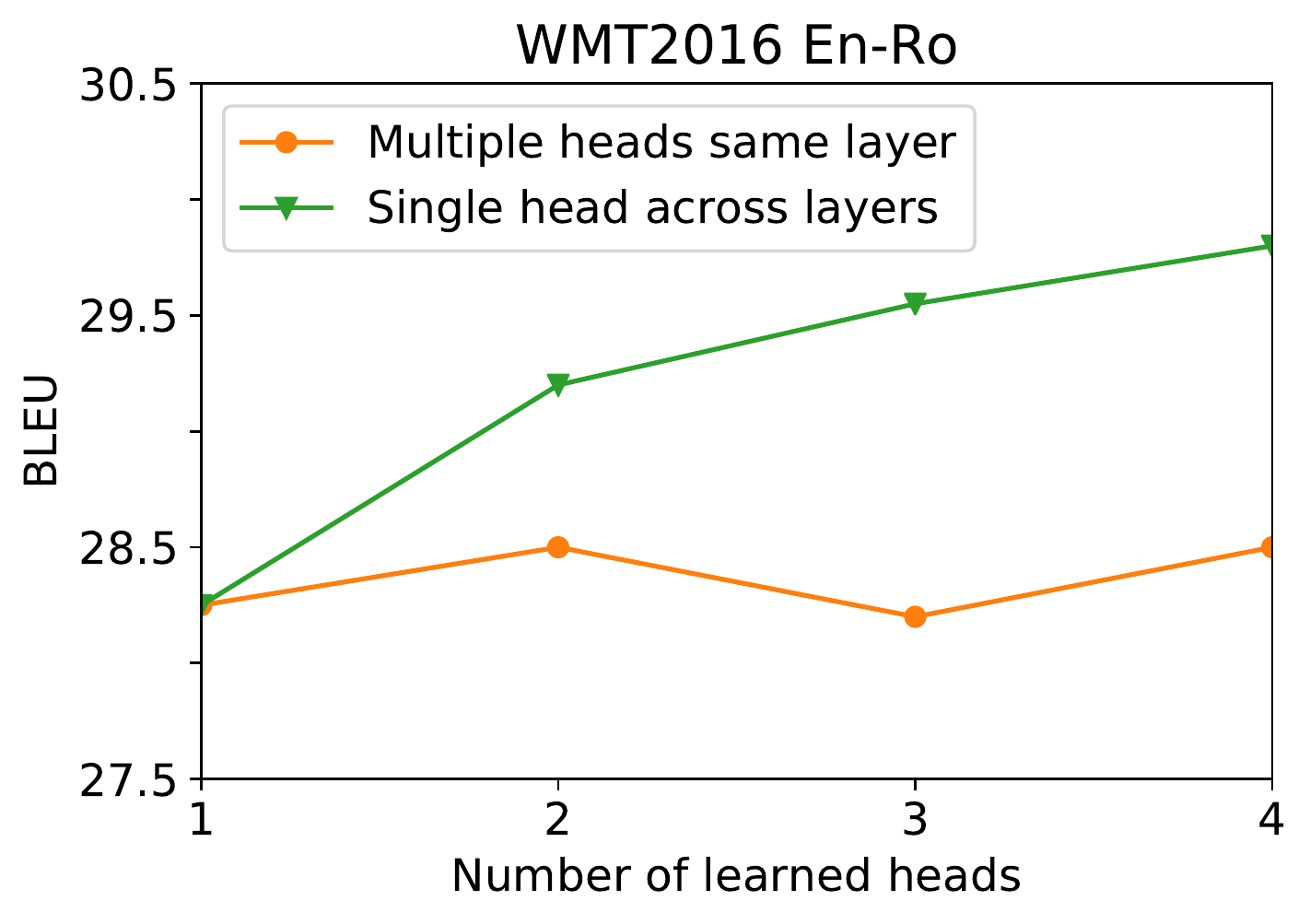}
    \caption{Adding more cross attention heads in the same layer helps less than adding individual heads across different layers.}
    \label{fig:enro}
\end{figure}

\paragraph{Where should we add additional cross attention heads?} 
Our experiments with cross attention so far have been limited to learning just a single head, as we have mainly been interested in minimal configurations. If we have a larger budget of cross attention heads, where should we put them? Is it better to have more cross attention heads in the last layer in the decoder (and no heads anywhere else), or to distribute them across multiple layers of the decoder? Experiments on the WMT16 En-Ro dataset\footnote{We used the smaller IWSLT En-De architecture for this experiment.} (Figure~\ref{fig:enro}) indicate that distributing learned heads over multiple layers leads to significantly better BLEU than adding all of them to the same layer. 


\section{Related Work}\label{sec:related}

Attention mechanisms were first introduced to augment vanilla recurrent models~\citep{kalchbrenner-blunsom-2013-recurrent-continuous,NIPS2014_5346,DBLP:journals/corr/BahdanauCB14,luong-etal-2015-effective,NIPS2015_5847,wu2016google,miceli-barone-etal-2017-deep} but have become the featured component of the state-of-the-art Transformer architecture~\citep{NIPS2017_7181} for NMT. We review recent research that focuses on analysing and improving multi-headed attention, and draw connections to our work.

The intuitive advantage of MHA is that different heads can focus on different types of information, all of which will eventually be helpful for translation.~\citet{voita-etal-2019-analyzing} find that some heads focus on adjacent tokens to the query (mirroring our analysis in Section~\ref{sec:background}), while others focus on specific dependency relations or rare tokens. ~\citet{correia-etal-2019-adaptively} discover that some heads are sensitive to subword clusters or interrogative words.~\citet{tang-etal-2018-self} shows that the number of MHA heads affects the ability to model long-range dependencies. ~\citet{NIPS2019_9551} show that pruning many heads from a pretrained model does not significantly impact BLEU scores. Similarly,  \citet{voita-etal-2019-analyzing} prune many encoder self-attention heads without degrading BLEU, while \citet{tang-etal-2019-understanding} further simplify the Transformer by removing the entire encoder for a drop of three BLEU points. In contrast to existing literature  on model pruning, we \textit{train} our models without learned attention heads instead of removing them post-hoc. 

There have been many efforts to modify MHA in Transformers. One such direction is to inject linguistic knowledge through auxiliary supervised tasks~\citep{garg-etal-2019-jointly,pham2019promoting}. Other work focuses on improving inference speed: ~\citet{yang-etal-2018-modeling} replace decoder self-attention with a simple average attention network, assigning equal weights to target-side previous tokens.\footnote{In preliminary experiments, we find that using uniform distributions for encoder self-attention decreases BLEU. This result is similar to the indexing implementation we describe in Section~\ref{subsec:modifications}.} ~\citet{wu2018pay} also speed up decoding by replacing self-attention with convolutions that have time-step dependent kernels; we further simplify this work with our fixed convolutional kernels in Section~\ref{sec:analysis}. \citet{cui-etal-2019-mixed} also explore fixed attention while retaining some learned parameters, and \citet{Vashishth2019AttentionIA} show that using uniform or random attention deteriorates performances on paired sentences tasks including machine translation. Other work has also explored modeling locality~\citep{shaw-etal-2018-self,yang-etal-2018-modeling}. 



\section{Conclusion}\label{sec:conclusion}
In this paper, we present  ``hard-coded'' Gaussian attention, which while lacking any learned parameters can rival multi-headed attention for neural machine translation. Our experiments suggest that encoder and decoder self-attention is not crucial for translation quality compared to cross attention. We further find that a model with hard-coded self-attention and just a single cross attention head performs slightly worse than a baseline Transformer. Our work provides a foundation for future work into simpler and more computationally efficient neural machine translation.

\section*{Acknowledgments}\label{sec:ack}
We thank the anonymous reviewers for their thoughtful comments, Omer Levy for general guidance and for suggesting some of our efficiency experiments, the UMass NLP group for helpful comments on earlier drafts, Nader Akoury for assisting with modifications to his Transformer codebase, and Kalpesh Krishna for advice on the structure of the paper. 

\clearpage

\appendix


\section{Mixed position for hard-coded self-attention works the best}

\begin{table}[htbp]
    \centering
    \begin{tabular}{ccc}
    \toprule
     Enc-Config  & Dec-Config &  BLEU\\
     \hline
      ($l$, $l$)   &  ($l$, $l$) & 27.4 \\
      ($l$, $l$)  &   ($c$, $c$) & 27.8 \\
      ($l$, $l$)  &   ($l$, $c$) & 28.1 \\
      ($l$, $r$)  &   ($l$, $c$) & \textbf{30.3} \\
    \bottomrule
    \end{tabular}
    \caption{Search for best hard-coded configuration for hard-coded self-attention. `$l$' stands for left, focusing on $i-1$, `$r$' for $i+1$ and `$c$' for $i$.  Middle layers are ($l$,$r$) for encoder and ($l$,$c$) for decoder. Each cell  shows settings we used in the lowest and highest layer.}
    \label{tab:hc-sa-mixed-config}
\end{table}

\section{Memory efficiency and inference speedups}
Table \ref{tab:mem-time-profile} summarizes the results of our profiling experiments on IWSLT16 En-De development set.
\begin{table}[!h]
    \centering
    \begin{tabular}{cccc} 
        \toprule
          Model & BLEU & sent/sec & tokens/batch \\ \hline
         \base & 30.0 & 43.1 & 14.1k \\
         \hcsa & 30.3 & 44.0 & 15.1k \\
         \shx & 28.1 & 50.1 & 16k \\
         \midrule
        \base /\nosa & 22.8 & 46.1 & 16.1k \\
        \shx /\nosa & 14.9 & 54.9 & 17k \\
        \bottomrule
    \end{tabular}
    \caption{{Decoding speedup (in terms of sentences per second) and memory improvements (max tokens per batch) on IWSLT16 En-De for a variety of models. The last two rows refer to \base\ and \shx\ configurations whose self-attention is completely removed. }}
    \label{tab:mem-time-profile}
\end{table}

\begin{figure*}[h!]
    \centering
    \includegraphics[width=0.33\textwidth]{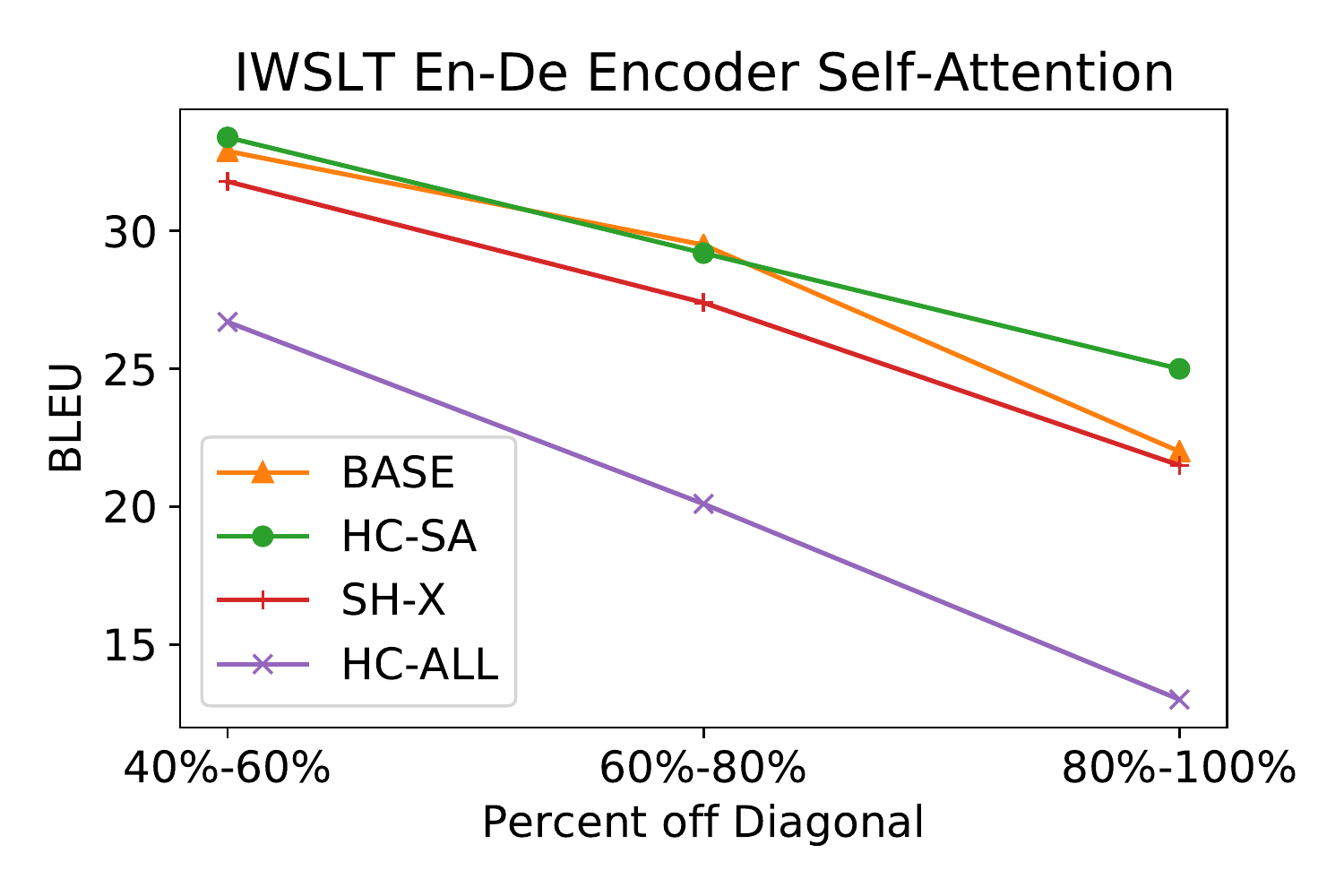}\includegraphics[width=0.33\textwidth]{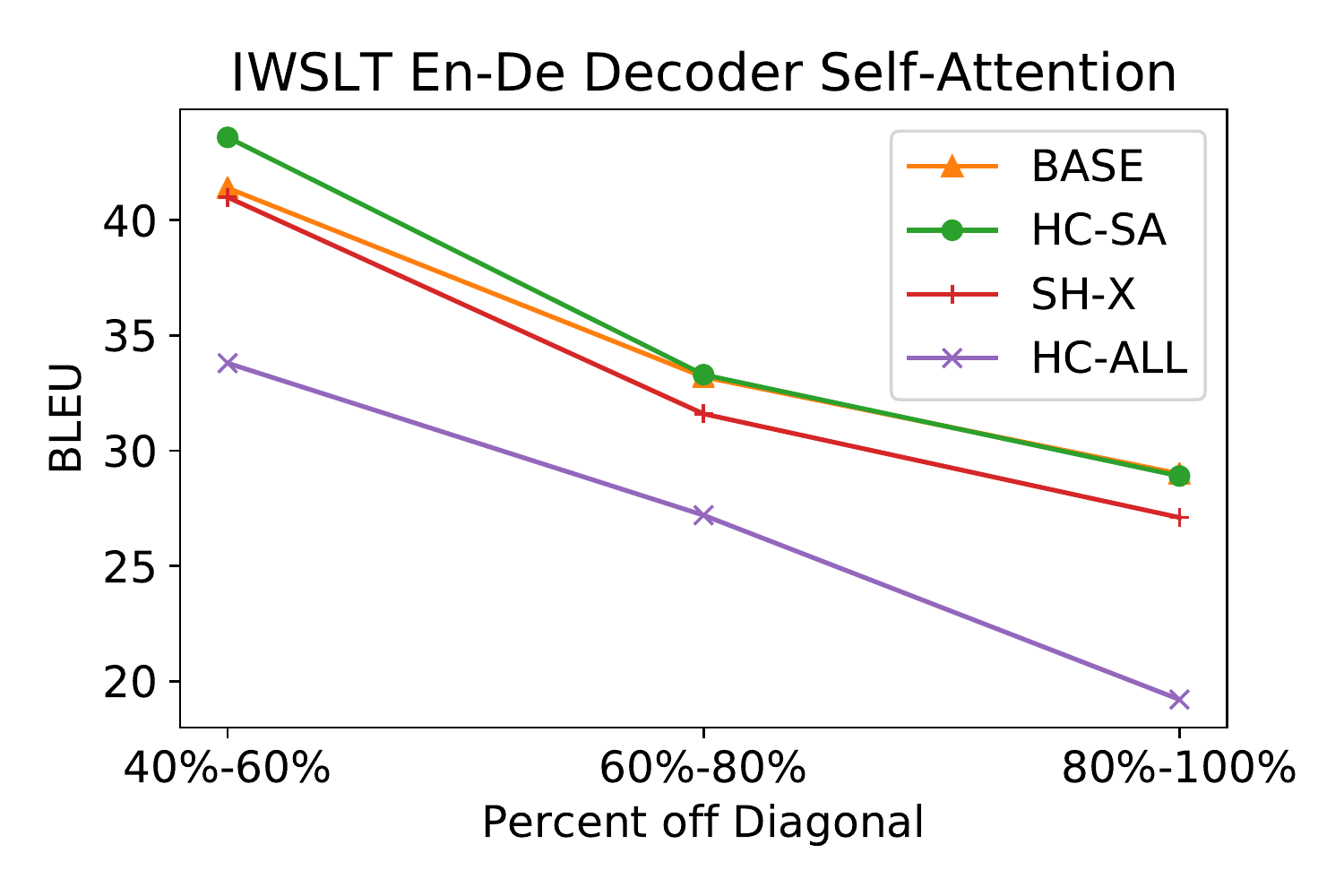} \includegraphics[width=0.33\textwidth]{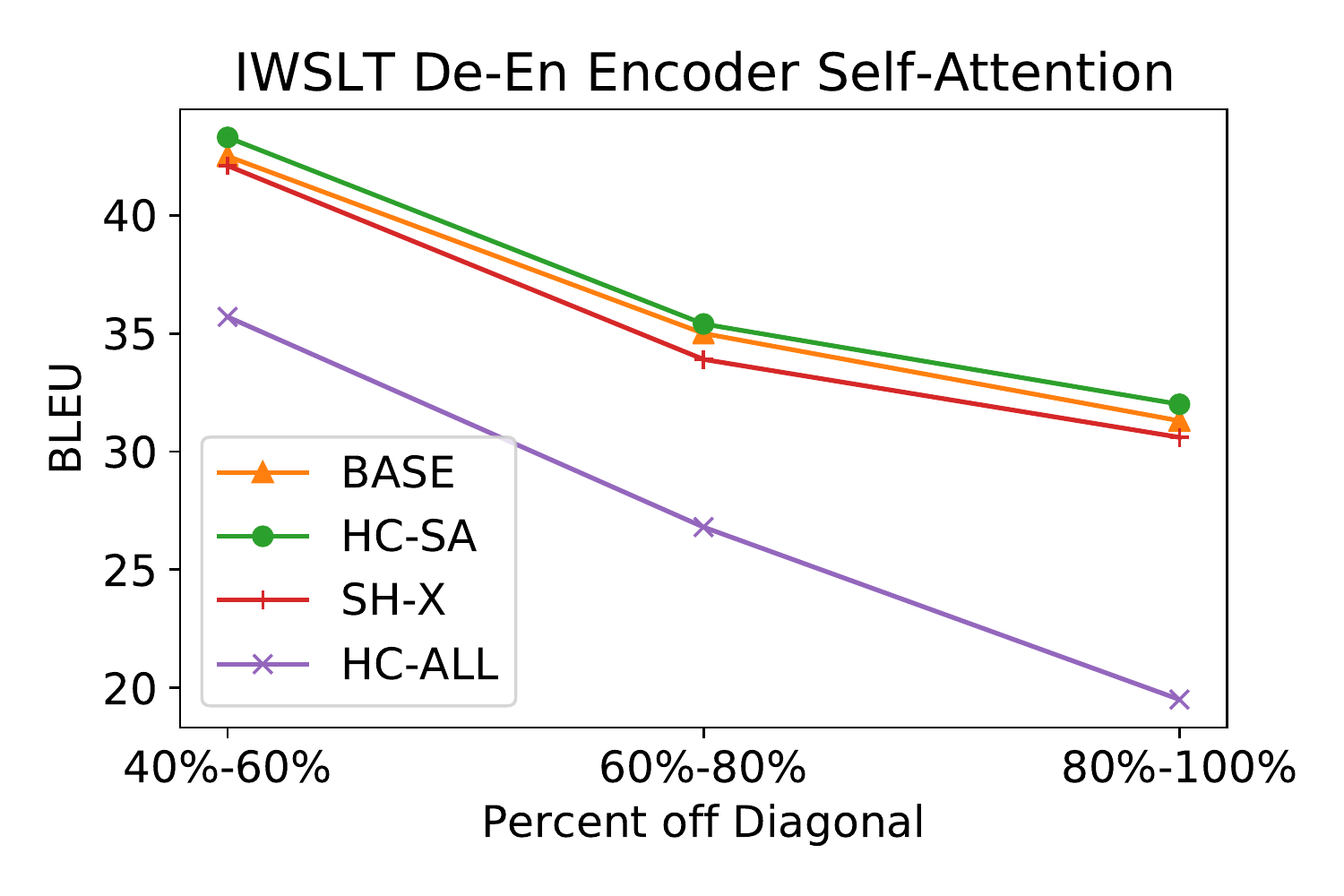}
    \caption{Off-diagonal analysis for IWSLT En-De/De-En self-attention}
    \label{fig:off-ende-analysis}
\end{figure*}

\section{Off-diagonal Analysis}
In addition to IWSLT16 De-En decoder self-attention analysis, we provide here the off-diagonal analysis results on IWSLT16 En-De encoder and decoder self-attention, and IWSLT16 De-En encoder self-attention in Figures~\ref{fig:off-ende-analysis}.

\end{document}